\documentclass[conference]{IEEEtran}

\usepackage{imakeidx}
\makeindex
\usepackage{cite}
\usepackage[dvipsnames]{xcolor}
\usepackage{url}
\usepackage{dutchcal}
\usepackage{graphicx}
\usepackage{multirow}
\graphicspath{{./images/}}
\usepackage{fancyhdr}

\begin{document}
\sloppy

% \title{Event-Level Emotion Recognition from Real-World RGB Photographs Using Deep Facial Expression Analysis}

\title{Event-Level Emotion Recognition in the Wild Using Deep Facial Expression Analysis}

\author{
    \IEEEauthorblockN{Aleksandr Semerikov}
    \IEEEauthorblockA{
        \textit{School of Information Technology and Engineering} \\
        \textit{Kazakh-British Technical University} \\
         Almaty, Kazakhstan\\
        \tt a\_semerikov@kbtu.kz
    }
    \and
    \IEEEauthorblockN{Pakizar Shamoi}
    \IEEEauthorblockA{
        \textit{School of Information Technology and Engineering} \\
        \textit{Kazakh-British Technical University} \\
         Almaty, Kazakhstan\\
        \tt p.shamoi@kbtu.kz
    }
}

\maketitle
\IEEEpeerreviewmaketitle
\fancypagestyle{firstpagestyle}{
    \fancyhf{} % clear all header and footer fields
    \fancyhead[L]{2026 IEEE 6th International Conference on Smart Information Systems and Technologies (SIST) \\ 13-15 May, 2026, Astana, Kazakhstan}
    \renewcommand{\headrulewidth}{0pt} % optional: remove header rule line
    \renewcommand{\footrulewidth}{0pt} % optional: remove footer rule line
}

\maketitle

\thispagestyle{firstpagestyle}

\begin{abstract}
Facial emotion recognition (FER) in real-world environments remains challenging due to unconstrained imaging conditions, including multiple faces, occlusions, pose variations, and complex lighting. Most existing studies focus on individual facial emotion classification and do not address the analysis of collective emotional states at the event level. This paper proposes an end-to-end pipeline for event-level emotion recognition from photographs. The approach detects faces in each image, classifies facial expressions using a deep convolutional neural network, and aggregates face-level emotion probabilities to estimate the overall emotional distribution of a public event. A comparative evaluation of several CNN architectures on the FER-2013 and RAF-DB datasets demonstrates that transfer learning with EfficientNet-B2 trained on RAF-DB is more suitable for real-world RGB data. The proposed method is evaluated on a real-world event dataset containing 1658 images. Experimental results show stable emotion distributions across event subsets, confirming the effectiveness of event-level aggregation for emotion analysis in the wild.
\end{abstract}
% =====================================================
\section{Introduction}

Facial emotion recognition (FER) is a fundamental task in computer vision and affective computing, aiming to identify human emotional states based on facial expressions \cite{Zhao2019FERSurvey}.
Automatic FER systems play an important role in a wide range of applications, including human-computer interaction, social behavior analysis, healthcare monitoring, marketing research, and sports analytics \cite{Jayaswal2025}.
In recent years, the growing availability of visual data from public events has increased interest in large-scale, real-world emotion analysis \cite{Leong2023Facial}.

% , FER remains a challenging problem in unconstrained environments. 
FER has achieved very high accuracy in lab settings. Despite significant progress in deep learning, it still struggles in unconstrained, real‑world conditions due to variability in faces, scenes, and capture conditions \cite{Karnati2023Understanding, Khan2022Facial, Dhall2022Emotion}.
Real-world photographs often contain multiple people, non-frontal faces, occlusions, motion blur, and highly variable lighting conditions \cite{Samadiani2019A, Li2019Reliable}.
Such factors significantly degrade the performance of models trained on controlled datasets.
As a result, many state-of-the-art FER systems demonstrate high accuracy on benchmarks but fail to generalize to practical deployment scenarios.

Most existing FER studies focus on single-face emotion classification. However, in applications such as event analysis, the overall emotional atmosphere is often more informative than individual predictions. A growing but still small line of research is now tackling group- or crowd-level emotion estimation, often by aggregating multiple faces \cite{Li2017RAFDB, Veltmeijer2023Automatic}.

This motivates the development of event-level emotion recognition systems that aggregate predictions across multiple images and faces.

In this work, we propose an end-to-end pipeline for estimating the emotional distribution of a public event using unconstrained RGB photographs.
The system detects faces, classifies facial expressions using a deep convolutional neural network, and aggregates predictions to obtain a robust event-level representation.
Special emphasis is placed on selecting a model well-suited to real-world RGB data and transfer learning.

1. The contributions of this study are as follows:
\begin{itemize}
    \item Comparative evaluation of CNN architectures and training strategies across FER-2013 and RAF-DB datasets for real-world facial emotion recognition.
    \item An aggregation strategy that stabilizes noisy face-level predictions and enables robust event-level emotion estimation.
    \item A real-world case study on a public sports event dataset.
\end{itemize}

Section I is this Introduction.  Section II reviews related work. Section III presents the proposed event-level emotion recognition pipeline, which includes dataset selection, model training, and an aggregation strategy. Section IV reports experimental results and comparative evaluations on FER-2013, RAF-DB, and a real-world event dataset. Section V provides a discussion. Finally, Section VI concludes the paper and outlines directions for future research.
% =====================================================
\section{Related Work}

\subsection{Classical FER}
% Early FER approaches relied on handcrafted features such as Local Binary Patterns (LBP), Gabor wavelets, and Histogram of Oriented Gradients (HOG).
% These features were typically combined with traditional classifiers such as Support Vector Machines or Random Forests.
% While such methods achieved reasonable performance under controlled conditions, they were highly sensitive to noise, pose variations, and illumination changes, which limited their applicability to real-world scenarios.
Early classical FER systems mainly used handcrafted texture/shape descriptors plus shallow classifiers.
Common 2D feature extractors include:
\begin{itemize}
    \item \textbf{Local Binary Patterns.} Encodes local texture. Widely used in global and local face descriptions, with good rotation and gray-scale invariance \cite{Li2025A, Niu2021Facial}.
    \item \textbf{Histogram of Oriented Gradients.} Captures gradient orientation. Robust to small geometric/optical deformations and widely applied to facial expression patterns \cite{Lakshmi2021Facial, Niu2021Facial, Yaddaden2021Facial}.
    \item \textbf{Gabor filters/wavelets.} Multi-scale, multi-orientation filters for facial texture and edges, often combined with other descriptors \cite{Niu2021Facial}.
\end{itemize}

These features were typically fed to SVM, k-NN, Random Forest, AdaBoost, HMM, and simple neural networks for multiclass emotion recognition \cite{Niu2021Facial}. Research consistently shows strong degradation of these systems under pose, illumination, and occlusion changes \cite{Nonis20193D, Samadiani2019A, Canedo2019Facial}.

\subsection{Deep Learning-Based FER}
Deep learning has transformed FER by replacing handcrafted features with end‑to‑end trainable models that better handle in‑the‑wild data. Deep CNNs (often with residual or Inception blocks) learn hierarchical facial features directly from pixels and consistently outperform classical models, especially on FER2013, CK+, JAFFE and in-the-wild sets
 \cite{Zhao2019FERSurvey, Minaee2019Deep-Emotion:, Jain2019Extended, Mollahosseini2015Going, Chowdary2021Deep, Meena2023Identifying}.

 Spatial/channel attention focuses on emotion-relevant regions (eyes, mouth, nose), improving robustness to occlusion and pose; examples include Deep-Emotion’s attentional CNN and attention-guided or MA-Net architectures \cite{Zhao2021Learning, Huang2023A, Saurav2024An}. Lightweight CNNs and transfer learning: Compact networks, squeeze-and-excitation ResNets, and pre-trained backbones (VGG, ResNet, MobileNet) adapt via fine-tuning to smaller FER datasets with strong accuracy and lower computation \cite{Shao2019Three, Pereira2024Systematic, Chowdary2021Deep, Grover2025Enhancing}.

% Several studies report that deeper networks either fail to converge or suffer from severe overfitting when trained on FER-2013.

The FER-2013 dataset became one of the earliest large-scale benchmarks for deep FER research \cite{Goodfellow2013FER}. Architectures such as VGG, ResNet, and Xception have been widely explored for FER tasks. However, its low-resolution grayscale images and noisy annotations limit the effectiveness of modern architectures and transfer learning techniques.

% \subsection{RAF-DB and Real-World FER}
 % \colorbox{Dandelion}{9. Add 5 papers}
To address the shortcomings of the FER-2013 dataset, the Real-world Affective Faces Database (RAF-DB) was introduced \cite{Li2017RAFDB}.
RAF-DB consists of higher-resolution RGB images collected from real-world sources and annotated using a crowdsourcing-based consensus strategy.
This dataset enables the training of models that generalize better to unconstrained environments.

Recent studies show that transfer learning from ImageNet-pretrained networks significantly improves FER performance on the RAF-DB dataset.
EfficientNet architectures, in particular, provide a favorable balance between accuracy and computational cost \cite{Tan2019}.

% \subsection{Face Detection in the Wild}
%  \colorbox{Dandelion}{10. Add 5 papers}
Accurate face detection is a critical component of real-world FER pipelines.
MediaPipe Face Detection offers a lightweight, efficient solution for large-scale image processing \cite{Lugaresi2019MediaPipe}. However, detecting small, partially occluded, or profile faces remains challenging and directly affects downstream emotion recognition performance.

% Overall, existing studies highlight the importance of dataset quality, model architecture selection, and robustness to unconstrained conditions, which motivates the approach proposed in this work.

Recent research highlights two persistent problems: overfitting due to limited labeled data and expression-unrelated variations (illumination, head pose, occlusion, identity bias) \cite{Karnati2023Understanding, Ge2022Facial, Li2018Deep, Pereira2024Systematic}.

% =====================================================
\section{Methods}

\subsection{Data}

\subsubsection{FER-2013}
\begin{figure}[tb]
\centering
\includegraphics[width=0.5\textwidth]{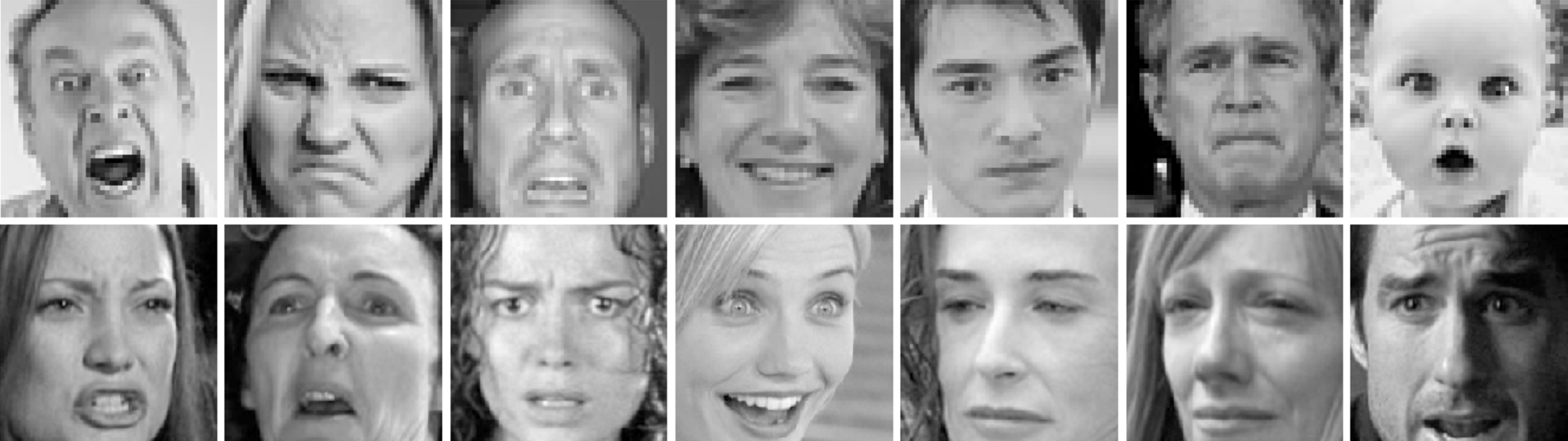}
\caption{Sample images from FER-2013 dataset.}
\label{fig:dataset-card}
\end{figure}

Initial experiments were conducted on the FER-2013 dataset \cite{Goodfellow2013FER}.
 The FER-2013 dataset is one of the most widely used benchmarks for facial emotion recognition. It was originally introduced as a part of the ICML 2013 Facial Expression Recognition Challenge. The dataset contains 35887 grayscale facial images, divided into a train set (28709 images), a public test set (3589 images), and a private test set (3589 images), each of size $48 \times 48$. All the images are labeled with 7 emotion categories: Angry, Disgusted, Fear, Happy, Neutral, Sad, and Surprise (see Fig.~\ref{fig:dataset-card}).

  \begin{figure}[tb]
\centering
\includegraphics[width=0.5\textwidth]{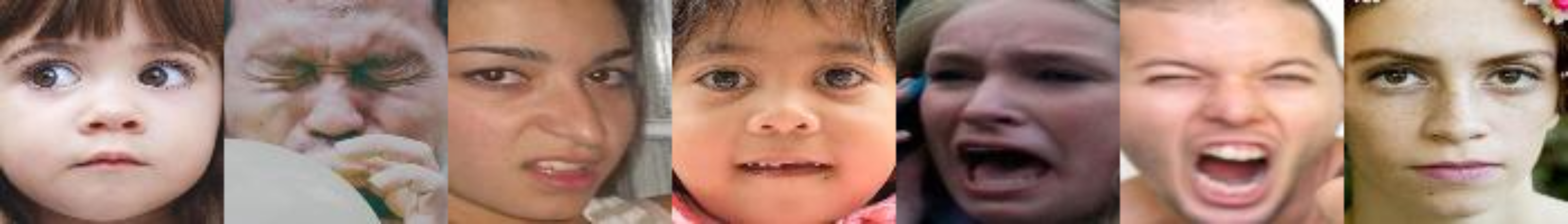}
\caption{Sample images from the RAF-DB dataset.}
\label{fig:raf-dataset-card}
\end{figure}
  
\begin{figure}[tb]
\centering
\includegraphics[width=\linewidth]{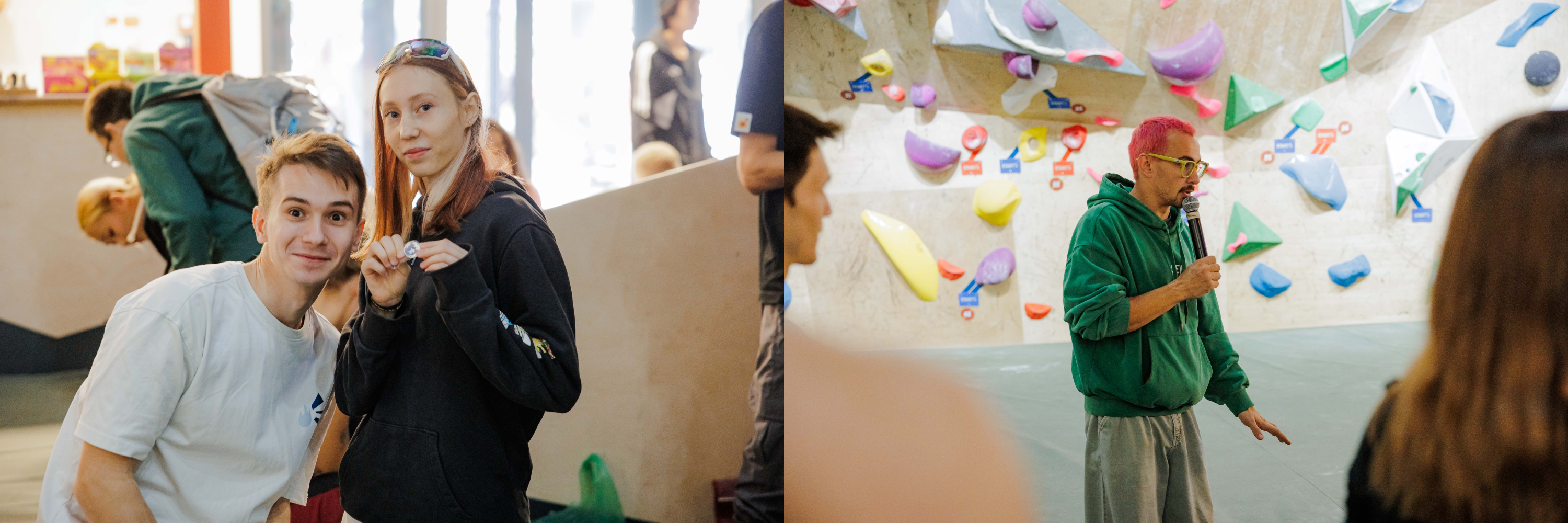}
\caption{Examples of Cave Fall Fest Dataset images.}
\label{fig:fest-dataset}
\end{figure}
\begin{figure*}[tb]
\centering
\includegraphics[width=\linewidth]{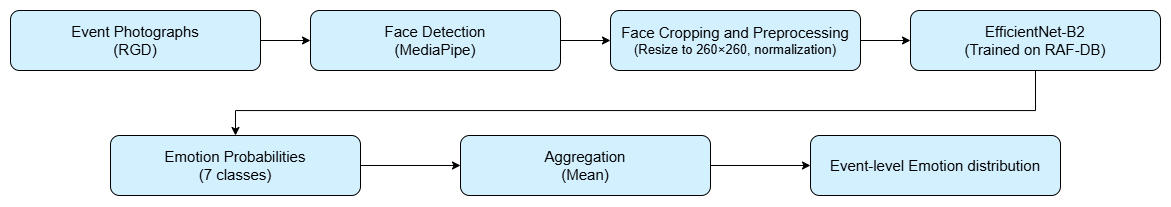}
\caption{Overview of the proposed event-level emotion recognition pipeline.}
\label{fig:pipeline}
\end{figure*}

\subsubsection{RAF-DB}

To overcome these limitations, further experiments were conducted on the RAF-DB dataset \cite{Li2017RAFDB}.
The RAF-DB (Real-World Affective Face Database) is a large-scale dataset of facial expressions collected from real-world images. It contains 15339 facial RGB images downloaded from the internet and manually annotated by 40 independent human coders. The final emotion label is obtained through a crowd-sourcing aggregation procedure. The RAF-DB was divided into a training set of 12271 images and a test set of 3068 images. The images are labeled with 7 basic emotions: Angry, Disgusted, Fear, Happy, Neutral, Sad, and Surprise (see Fig.~\ref{fig:raf-dataset-card}).
Images differ in subjects' age, gender, and ethnicity, head poses, lighting conditions, and occlusions (glasses, facial hair, or self-occlusion).

\subsubsection{Cave Fall Fest Dataset}

For applied evaluation, a proprietary dataset of 1658 RGB photographs was collected during the Cave Fall Fest climbing event.
Cave Fall Fest was a climbing event held on 18 October 2025. It was organized into 4 climbing sets, each corresponding to a distinct time interval. The dataset includes separate images for each climbing set, each containing multiple human faces, a single face, or no faces at all, with the number of detectable faces varying from image to image (Fig.~\ref{fig:fest-dataset}).
The dataset reflects real-world conditions, including multiple faces per image, profile views, partial occlusions, and varying lighting conditions. It makes the dataset representative of naturalistic conditions encountered during large indoor sporting events and well-suited for evaluating the robustness of face detection and emotion recognition models.

\subsection{Proposed Event-Level Emotion Recognition Pipeline}

The proposed pipeline consists of sequential stages designed to handle unconstrained real-world photographs(see Fig.~\ref{fig:pipeline}).
Faces are detected in each image using the MediaPipe Face Detection framework.
Detected faces are cropped and normalized to match the input requirements of the emotion recognition model.

Facial emotion classification is performed using an EfficientNet-B2 convolutional neural network trained on the RAF-DB dataset. Images without faces were not included in the calculation of the mean probability. 
For each detected face, the trained EfficientNet-B2 model
produces a probability distribution over seven basic emotions
classes:
\begin{equation}
\mathbf{p}_i =
[p_i^{(1)}, p_i^{(2)}, \dots, p_i^{(7)}],
\end{equation}
corresponding to the emotions \textit{surprised, fear, disgust,
happy, sad, angry, and neutral}. Since the final layer of the network uses a softmax activation function, the predicted
probabilities satisfy the normalization constraint:
\begin{equation}
\sum_{k=1}^{7} p_i^{(k)} = 1.
\end{equation}
Thus, instead of assigning a single discrete label, the model
provides a confidence-weighted representation of emotional
expression for each face.
To obtain event-level statistics, images were grouped into four subsets corresponding to different climbing
sets. Let $M_s$ denote the total number of detected faces
within subset $s$. The mean probability of emotion class $k$
for subset $s$ was computed as:
\begin{equation}
\bar{p}_s^{(k)} =
\frac{1}{M_s}
\sum_{i=1}^{M_s} p_i^{(k)}.
\end{equation}
This aggregation strategy produces a 7-dimensional
mean probability vector for each subset:
\begin{equation}
\bar{\mathbf{p}}_s =
[\bar{p}_s^{(1)}, \bar{p}_s^{(2)}, \dots, \bar{p}_s^{(7)}].
\end{equation}
Averaging softmax probabilities across all detected faces
reduces the influence of individual misclassifications and
provides a stable probabilistic characterization of the overall
emotional distribution within each event segment.

% =====================================================
\section{Results}

\subsection{Comparative Evaluation}

Several architectures were evaluated, including Mini-Xception \cite{Arriaga2017MiniXception}, a custom CNN, and transfer learning models based on MobileNetV2 and EfficientNet-B2.

Table~\ref{tab:cnn_comparison} compares the performance of different CNN architectures on the FER-2013 and RAF-DB datasets. On FER-2013, lightweight models specifically designed for low-resolution grayscale images, such as Mini-Xception and a custom CNN, outperform transfer-learning approaches. In contrast, ImageNet-pretrained models exhibit poor performance due to the mismatch between FER-2013’s grayscale, low-quality data distribution, and RGB feature representations learned during pretraining.

On RAF-DB, which consists of higher-resolution RGB images collected under real-world conditions, transfer learning models demonstrate clear advantages. In particular, EfficientNet-B2 achieves the highest test accuracy, confirming its suitability for unconstrained facial emotion recognition. 

% These results highlight the critical role of dataset characteristics in model selection and motivate the use of EfficientNet-B2 for subsequent event-level emotion analysis in the wild.

% \begin{table}[tb]
% \caption{Comparison of CNN architectures on FER-2013 and RAF-DB datasets}
% \centering
% \begin{tabular}{lccccc}
% \hline
% Dataset & Model & Input size & Params (M) & Test accuracy \\
% \hline
% \multirow{4}{*}{FER-2013} 
% & Mini-Xception & $48 \times 48 \times 1$ & 1.54 & \textbf{0.656} \\
% & Custom CNN & $48 \times 48 \times 1$ & 1.28 & 0.632 \\
% & MobileNetV2 (TL) & $128 \times 128 \times 3$ & 2.27 & 0.247 \\
% & EfficientNet-B2 (TL) & $224 \times 224 \times 3$ & 7.78 & 0.247 \\
% \hline
% \multirow{4}{*}{RAF-DB} 
% & Mini-Xception & $96 \times 96 \times 3$ & 1.54 & 0.684 \\
% & MobileNetV2 (TL) & $128 \times 128 \times 3$ & 2.27 & 0.701 \\
% & Custom CNN & $96 \times 96 \times 3$ & 0.48 & 0.721 \\
% & EfficientNet-B2 (TL) & $260 \times 260 \times 3$ & 7.78 & \textbf{0.750} \\
% \hline
% \end{tabular}
% \label{tab:cnn_comparison}
% \end{table}
\begin{table*}[!htbp]
\caption{Comparison of CNN architectures on FER-2013 and RAF-DB datasets}
\centering
\begin{tabular}{lccccc}
\hline
Dataset & Model & Input size & Params (M) & Test accuracy \\
\hline
\multirow{4}{*}{FER-2013} 
& Mini-Xception & $48 \times 48 \times 1$ & 1.54 & \textbf{0.656} \\
& Custom CNN & $48 \times 48 \times 1$ & 1.28 & 0.632 \\
& MobileNetV2 (TL) & $128 \times 128 \times 3$ & 2.27 & 0.247 \\
& EfficientNet-B2 (TL) & $224 \times 224 \times 3$ & 7.78 & 0.247 \\
\hline
\multirow{4}{*}{RAF-DB} 
& Mini-Xception & $96 \times 96 \times 3$ & 1.54 & 0.684 \\
& MobileNetV2 (TL) & $128 \times 128 \times 3$ & 2.27 & 0.701 \\
& Custom CNN & $96 \times 96 \times 3$ & 0.48 & 0.721 \\
& EfficientNet-B2 (TL) & $260 \times 260 \times 3$ & 7.78 & \textbf{0.750} \\
\hline
\end{tabular}
\label{tab:cnn_comparison}
\end{table*}

\begin{table*}[ht]
\centering
\caption{Mean Emotion Probabilities Across the Climbing Sets}
\label{tab:emotion_summary}
\begin{tabular}{lccccccc}
\hline
\textbf{Set} & \textbf{Surprised} & \textbf{Fear} & \textbf{Disgust} & \textbf{Happy} & \textbf{Sad} & \textbf{Angry} & \textbf{Neutral} \\
\hline
1 -- Set & 0.180 & 0.007 & 0.032 & 0.085 & 0.167 & 0.014 & 0.514 \\
2 -- Set & 0.203 & 0.003 & 0.021 & 0.020 & 0.162 & 0.007 & 0.584 \\
3 -- Set & 0.211 & 0.006 & 0.036 & 0.071 & 0.158 & 0.015 & 0.503 \\
4 -- Set & 0.175 & 0.006 & 0.023 & 0.057 & 0.153 & 0.017 & 0.570 \\
\hline
\end{tabular}
\end{table*}

% \begin{table}[htbp]
% \caption{Comparison of CNN architectures on the FER-2013 dataset}
% \centering
% \begin{tabular}{lcccc}
% \hline
% Model & Input size & Params (M) & Test accuracy \\
% \hline
% Mini-Xception & $48 \times 48 \times 1$ & 1.54 & \textbf{0.656} \\
% Custom CNN & $48 \times 48 \times 1$ & 1.28 & 0.632 \\
% MobileNetV2 (TL) & $128 \times 128 \times 3$ & 2.27 & 0.247 \\
% EfficientNet-B2 (TL) & $224 \times 224 \times 3$ & 7.78 & 0.247 \\
% \hline
% \end{tabular}
% \label{tab:fer2013_results}
% \end{table}

% As shown in Table~I, lightweight architectures designed for grayscale images outperform transfer learning models on FER-2013.
% This behavior can be explained by the mismatch between the FER-2013 data distribution and ImageNet-pretrained RGB feature representations.

% \begin{table}[htbp]
% \caption{Comparison of CNN architectures on the RAF-DB dataset}
% \centering
% \begin{tabular}{lcccc}
% \hline
% Model & Input size & Params (M) & Test accuracy \\
% \hline
% Mini-Xception & $96 \times 96 \times 3$ & 1.54 & 0.684 \\
% MobileNetV2 (TL) & $128 \times 128 \times 3$ & 2.27 & 0.701 \\
% Custom CNN & $96 \times 96 \times 3$ & 0.48 & 0.721 \\
% EfficientNet-B2 (TL) & $260 \times 260 \times 3$ & 7.78 & \textbf{0.750} \\
% \hline
% \end{tabular}
% \label{tab:rafdb_results}
% \end{table}

% Table~II shows that EfficientNet-B2 achieves the highest accuracy on RAF-DB, confirming its suitability for real-world RGB facial emotion recognition.

% \subsection{Quantitative Evaluation}

 \subsection{Case study on a public sports event dataset}
The EfficientNet-B2 model, trained on the RAF-DB dataset, was applied to the Cave Fall Fest dataset.
Images without detected faces were excluded to reduce noise.

\begin{figure*}[htbp]
\centering
\includegraphics[width=\linewidth]{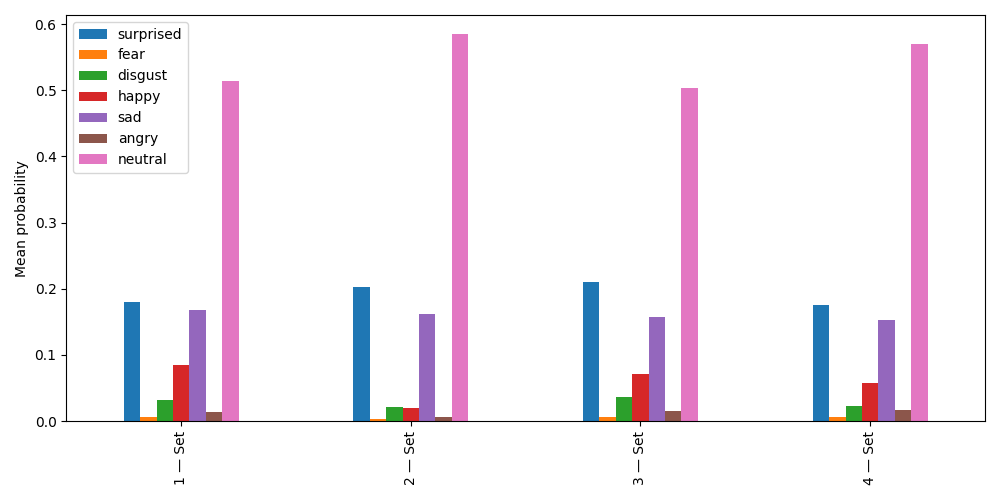}
\caption{Aggregated emotion distributions across four event sets.}
\label{fig:event_distribution}
\end{figure*}

Neutral and happy emotions dominated across all subsets (see Fig.~\ref{fig:event_distribution}).
The consistency of emotion distributions across subsets indicates robustness of the proposed aggregation strategy. Table II summarizes the mean emotion probabilities across the four climbing sets. The neutral class clearly dominates, with values ranging from 0.503 to 0.584. This outcome is expected, as the neutral label is typically assigned when the model is uncertain or when the facial expression lacks distinctive emotional cues.
Among the remaining categories, surprise and sadness show the highest mean probabilities, while fear, disgust, anger, and happiness remain consistently low across all sets. Overall, the distribution reflects the natural conditions of the climbing event, in which facial expressions are subtle, short-lived, and often affected by motion, occlusion, or non-frontal views.

% \subsection{Qualitative Analysis}

\begin{figure}[htbp]
\centering
\includegraphics[width=\linewidth]{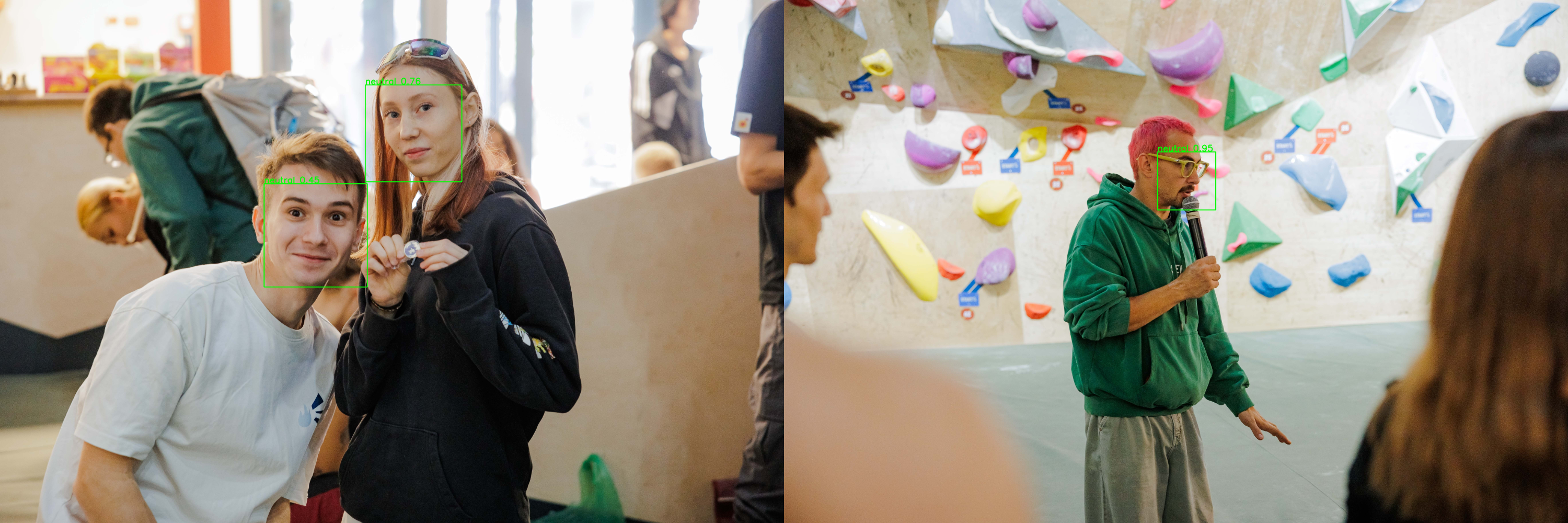}
\caption{Examples of detected faces with predicted emotions and confidence scores.}
\label{fig:bbox_examples}
\end{figure}

Visual inspection of annotated images (Fig.~\ref{fig:bbox_examples}) confirms that the system successfully detects multiple faces and produces plausible emotion predictions.
Failures mainly occur for very small faces, extreme head poses, and heavy occlusions.

\begin{figure}[htbp]
\centering
\includegraphics[width=\linewidth]{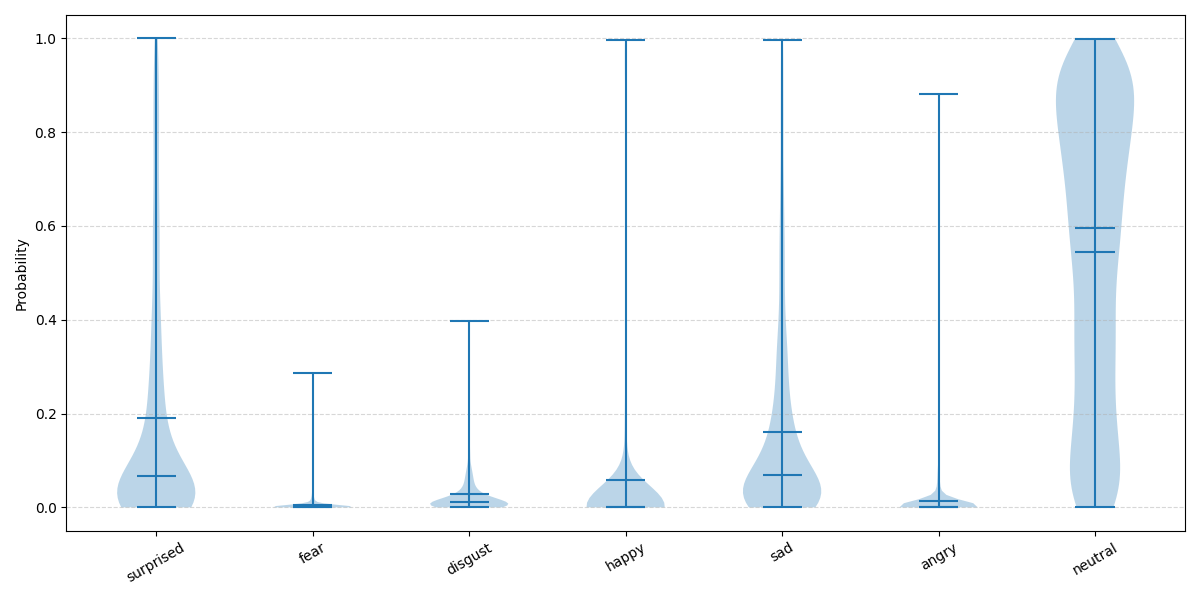}
\caption{Violin plot of the emotion distribution of the faces in Cave Fall Fest dataset.}
\label{fig:violin}
\end{figure}

\begin{figure}[ht]
\centering
\includegraphics[width=\linewidth]{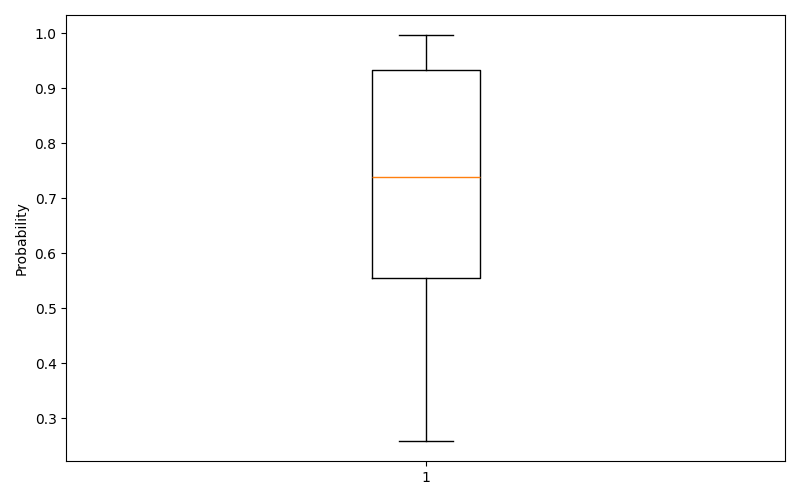}
\caption{Confidence distribution for happy class.}
\label{fig:boxplot}
\end{figure}

The violin plot (Fig.~\ref{fig:violin}) reveals highly skewed and competitive probability distributions induced by the softmax function. Most class probabilities are concentrated near zero, while only a small subset of samples achieves high confidence. The neutral class exhibits the broadest and highest-density distribution, confirming its dominance, whereas other emotions display sparse but distinct high-probability tails, indicating occasional confident detections.
The confidence distribution for the happy class(Fig.~\ref{fig:boxplot}), conditioned on it being the dominant prediction, demonstrates that although happy is not the most frequent class, it is typically predicted with relatively high confidence. This suggests that positive expressions are less common but visually distinctive, allowing the model to recognize them reliably when present.

The experiments highlight the critical role of dataset selection in FER tasks.
FER-2013 is suitable for benchmarking compact grayscale models but unsuitable for real-world RGB applications.
RAF-DB enables effective transfer learning and improved generalization.

% =====================================================
\section{Discussion}

      % \colorbox{Dandelion}{28. Comparison with recent studies 23-26 }
      %   \colorbox{Dandelion}{29. Find 3-4 citations }
      Let us examine how our findings corroborate existing studies employing automated or observer-based facial coding methods in challenging outdoor activities and physically demanding tasks.

Recent studies find neutral as the dominant facial category, with relatively low fear, anger, and disgust:

\begin{itemize}
    \item In preschoolers doing outdoor activities, neutral facial expressions were initially high and significantly decreased after intervention, indicating more non‑neutral expressions but not large spikes in fear/anger/disgust \cite{Boobani2025Utilizing}.

    \item In downhill mountain biking, online facial data showed substantial neutral and mixed low‑intensity expressions, with some surprise and anger‑like patterns; strong basic emotions were not dominant moment‑to‑moment \cite{Hetland2019The}.

    \item Studies on body and face emotion sets show that neutral is often the most accurately recognized category, followed by sadness and some other emotions, with fear and disgust typically recognized less reliably \cite{Zhang2025Multidimensional, Calvo2016Recognition}.
    
\end{itemize}

These patterns are consistent with our findings: a dominance of neutral, followed by emotions like surprise and sadness, while fear, disgust, and anger remain low.

One explanation for the prevalence of neutral emotions is that concentration can be misclassified as neutral or mild anger. Climbers under load may show “concentration faces” that tools label as neutral or weak negative emotion \cite{Hetland2019The}.

% Event-level aggregation significantly stabilizes predictions and reduces the influence of individual classification errors.
As the results show, people rarely exhibit fear during climbing sessions. That can be a good motivator for beginner climbers to overcome a lack of self-confidence and fear of engaging in new activities. Anger and sadness could be interpreted as disappointment resulting from the failure. That is the usual case for each person in each activity.
% Future work may explore more robust face detectors, temporal modeling of emotions across event timelines, and fuzzified results of emotion presence on each face.

% =====================================================
\section{Conclusion}
  % \colorbox{Dandelion}{25. More details from the results }
  %   \colorbox{Dandelion}{26. Limitations }
  %     \colorbox{Dandelion}{27. Future works }
This work presented an event-level facial emotion recognition pipeline tailored for real-world RGB photographs.
Through extensive experiments on FER-2013 and RAF-DB, EfficientNet-B2 trained on RAF-DB was selected as the most suitable architecture.  

The proposed system was successfully applied to a real-world event dataset, demonstrating its practical applicability. The aggregated emotion distributions across the four event subsets show a consistent dominance of the neutral class, followed by surprised and sad. At the same time, fear, disgust, and anger remain relatively unlikely. Our findings suggest a generally neutral, mildly positive emotional atmosphere throughout the event. Event-level aggregation significantly stabilizes predictions and reduces the influence of individual classification errors.

As for limitations, our study relies on a single real-world event dataset without ground-truth event-level emotion annotations, which limits quantitative validation of the aggregated results. 
Future research will explore confidence-weighted aggregation strategies to improve reliability under challenging conditions.

\bibliography{library}

\end{document}